\documentclass[10pt,twocolumn,letterpaper]{article}

\usepackage{iccv}
\usepackage{times}
\usepackage{epsfig}
\usepackage{graphicx}
\usepackage{amsmath}
\usepackage{amssymb}
\usepackage{booktabs}
\usepackage{multirow}

\usepackage[pagebackref=true,breaklinks=true,letterpaper=true,colorlinks,bookmarks=false]{hyperref}

\iccvfinalcopy 

\def\iccvPaperID{8659} 

\ificcvfinal\fi

\begin{document}

\title{OPT: Omni-Perception Pre-Trainer for Cross-Modal Understanding and Generation}

\author{Jing Liu, Xinxin Zhu, Fei Liu, Longteng Guo, Zijia Zhao, Mingzhen Sun,   \\ 
Hanqing Lu, Weining Wang, Hanqing Lu, Shiyu Zhou, Jiajun Zhang, Jinqiao Wang  \\
Institute of Automation, Chinese Academy of Sciences \\
{\tt\small \{jliu, xinxin.zhu, longteng.guo, luhq, weining.wang, luhq, jjzhang, jqwang\}@nlpr.ia.ac.cn} \\
{\tt\small \{liufei2017, sunmingzhen2020, zhoushiyu2013\}@ia.ac.cn, zhaozj17@126.com}
}

\maketitle
\ificcvfinal\thispagestyle{empty}\fi

\renewcommand{\thefootnote}{}
\begin{abstract}
In this paper, we propose an Omni-perception Pre-Trainer (OPT) for cross-modal understanding and generation, by jointly modeling visual, text and audio resources. OPT is constructed in an encoder-decoder framework, including three single-modal encoders to generate token-based embeddings for each modality, a cross-modal encoder to encode the correlations among the three modalities, and two cross-modal decoders to generate text and image respectively. For the OPT's pre-training, we design a multi-task pretext learning scheme to model multi-modal resources from three different data granularities, \ie, token-, modality-, and sample-level modeling, through which OPT learns to align and translate among different modalities. The pre-training task is carried out on a large amount of image-text-audio triplets from Open Images. Experimental results show that OPT can learn strong image-text-audio multi-modal representations and achieve promising results on a variety of cross-modal understanding and generation tasks.
\end{abstract}

\section{Introduction}
Human can learn knowledge by reading, seeing, and hearing things, \ie, exploring multi-modal resources including text, vision (image or video), and audio. And they further utilize the learned knowledge to understand and interact with the world around them. Therefore, a machine with human-like intelligence should be trained on multi-modal resources, to develop the both capabilities of cross-modal understanding and generation.

Recently, the researches on various cross-modal applications have been widely concerned, \eg multimodal retrieval \cite{wang2017adversarial,gordo2017end}, speech recognition enhanced with video \cite{ghorbani2020,palaskar2018}, visual question answering \cite{antol2015,yu2019deep}, visual captioning \cite{you2016image,mao2014deep}, speech to image generation \cite{wang20s2igan,li20direct}, and text-to-image generation \cite{reed2016,xu2018attngan}, etc. However, they are proposed to specialize in certain types of cross-modal understanding or generation tasks, and cannot establish general knowledge for unified processing. Fortunately, recent advances on pre-trained models have pointed out a promising direction towards such human-like intelligent systems with \textit{pretrain-then-transfer} learning paradigm. Pre-trained models have made great processes in computer vision (CV) \cite{he2016deep,densenet}, natural language processing (NLP) \cite{bert,xlnet}, and speech processing \cite{dahl2011}. It is a pity that existing pre-trained models are mostly limited to single-modal tasks or fusing only two modalities (\eg, vision-and-text pretraining). So far, no pretrained model is designed to connect the three most common modalities, \ie, text, vision, and audio. In addition, rare pre-trained models have the both capabilities for cross-modal understanding and generation. Most existing multi-modal pretrained models are either developed for understanding tasks, \eg, ViLBERT \cite{lvilbert}, VisualBERT \cite{visualbert}, or restricted to generation tasks, \eg, DALL-E \cite{dalle}. We are therefore interested in connecting the domains of visual data, natural language utterances, and audio data to developing a unified pre-trained model for cross-modal understanding and generation.

In this paper, we propose an Omni-perception Pre-Trainer (OPT) by exploring text, visual, and audio resources, which is competent for cross modal understanding and generation in a unified framework. OPT is pre-trained on large amounts of language-vision-audio triplets with a multi-task pretext learning scheme, and can effectively adapt to downstream understanding and generation tasks given single-, two-, or three-modal inputs. The architecture of OPT has three kinds of components, including three single-modal encoders, a cross-modal encoder, and two cross-modal decoders. Specially, we first encode the image, text, and audio separately by three single modal encoders. The results of the encoders are three sequences of token embeddings, where each token represents a region in the image, a word in the text, and an audio clip in the audio. We then learn joint contextualized representations for these three modalities through a Transformer based cross-modal encoder, which provides interaction among modalities at different representation depths. Finally, two cross-modal decoders take the outputs of the cross-modal encoder to generate text and image respectively in autoregressive manners.

Cross-modal understanding requires aligning inputs from different modalities at both the fine-grained token-level and the coarse-grained sample-level. The capacity to translate among modalities, \ie cross-modal generation, is endowed with modality-level modeling. Therefore, we pre-train OPT on Open Images with three levels of pretext tasks: token-, modality-, and sample-level modeling. Token-level modeling predicts the semantics of masked tokens given the unmasked inputs. Modality-level modeling includes two generative tasks, \ie, denoising text reconstruction and denoising image reconstruction, and a novel modality-level masking mechanism. The modality-level masking mechanism randomly masks out the whole inputs from any one or two of the three modalities, allowing OPT to adapt to different downstream tasks with single-, two-, or three-modal inputs. Sample-level modeling learns the alignment among the three modalities corresponding to the same sample.

We extensively validate OPT on a number of downstream tasks, including cross-modal retrieval, multi-modal classification, visual question answering, cross-modal text generation (including speech recognition and visual captioning), and text-to-image generation. Experimental results demonstrate the effectiveness of OPT in comparison with some baselines. It is noted that our OPT achieved amazing performance on cross-model text generation tasks, which can further verify the advantages of our unified pre-training architecture, since such tasks require first to understand one modality correctly, and then to generate another modality with the similar semantics. 

Our OPT model mainly has the following advantages compared with previous pre-trained models:
\begin{itemize}
  \item OPT is the first pre-trained model that connects the three modalities of text, vision, and audio, and is endowed with the both capacities of cross-modal understanding and generation. 
  \item OPT learns to align and translate among different modalities with the token-, modality-, and sample-level pretext tasks.
  \item OPT can effectively adapt to and perform competitively on a series of cross-modal understanding and generation downstream tasks with parial or all modalities as inputs. 
\end{itemize}

\section{Related Work}
\textbf{Single-Modal Pre-Training.} Recently, self-supervised pre-trained language models, such as GPT \cite{gpt}, BERT \cite{bert}, XLNet \cite{xlnet}, MASS \cite{mass}, UniLM \cite{dong19unified} and BART \cite{bart}, have achieve great success on NLP tasks. GPT \cite{gpt} is one of the early successes, which exploits the unidirectional word context to learn general language representations. BERT \cite{bert} enables the learning of bidirectional representations with MLM (masked language modeling) and NSP (next sentence prediction) as proxy tasks. XLNet \cite{xlnet} improves BERT with a generalized autoregressive pretraining mechanism. These BERT-type pre-training models only support language understanding via one encoder. Several recent works go beyond the traditional language pre-training and propose encoder-decoder networks for language generation through generative proxy tasks (\eg, masked sequence to sequence learning in MASS \cite{mass}, sequence-to-sequence modeling in UniLM \cite{dong19unified}, and denoising sequence-to-sequence modeling in BART \cite{bart}). The keys to their success are the use of Transformer \cite{vaswani2017attention} architecture for learning contextualized representations and effective pre-training tasks over large-scale language corpus. In the field of computer vision, self-supervised pre-training of visual representation has made great progress, facilitating many downstream tasks such as image classification, object detection, and semantic segmentation. Previous methods focus on designing different pretext tasks. One of the most promising directions among them is contrastive learning \cite{oord2018representation}, which transforms one image into multiple views, and minimize the distance between views from the same image and maximize the distance between views from different images. Representative methods include SimCLR \cite{chen2020simple}, MoCo \cite{he2020momentum}, BYOL \cite{grill2020bootstrap}.
In audio and speech processing, pre-training has focused on emotion recognition \cite{lian2018improving}, speaker identification \cite{ravanelli2018learning}, phoneme discrimination \cite{synnaeve2016temporal,oord2018representation}, transferring ASR representations from one language to another \cite{kunze2017transfer},  unsupervised representations learning for speech \cite{schneider2019wav2vec}, audio representation learning \cite{wang2021multi}. 

\begin{figure*}[t]
\begin{center}
   \includegraphics[width=0.93\linewidth]{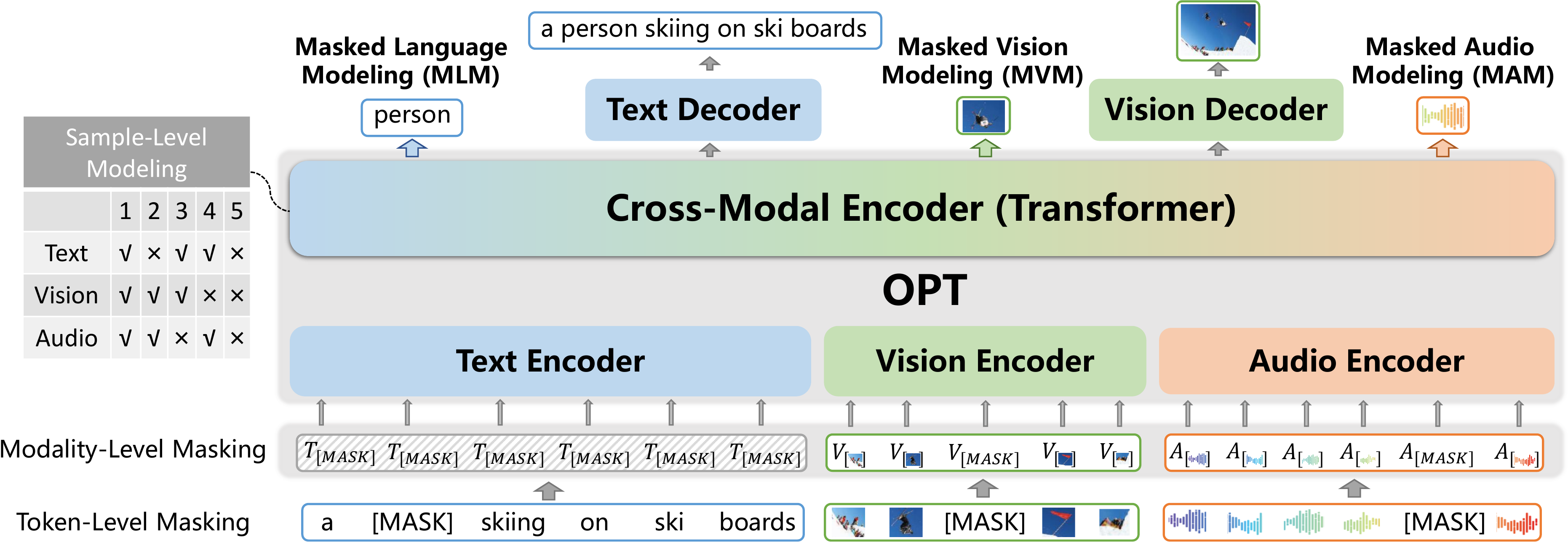}
\end{center}
  \caption{Model architecture of the proposed OPT, consisting of three single-modal encoders, a cross-modal encoder and two cross-modal decoders. We propose three levels of pre-training tasks: (1) token-level modeling, including masked language modeling (MLM), masked vision modeling (MVM), and masked audio modeling (MAM); (2) modality-level modeling, including denoising text reconstruction and denoising image reconstruction; and (3) sample-level modeling, where ``$\surd$'' denotes the corresponding modalities are matching. We introduce two masking mechanisms: (1) token-level masking, in order for token-level modeling; and (2) modality-level masking, in order for modality-level modeling and enabling arbitrary number of input modalities.}
\label{fig:model}
\end{figure*}

\textbf{Multi-Modal Pre-Training.} Inspired by language pre-training, the researchers starts to focus more on Vision+Language (VL) pre-training in multi-modal scenario. Current VL pretrained models can be divided into two types: one-stream and two-stream. One-stream approaches include VisualBERT \cite{visualbert}, UNITER \cite{chen2019uniter}, Unicoder-VL \cite{unicoder} and VL-BERT \cite{su2019vl}. VisualBERT \cite{visualbert} directly adapts BERT to VL pre-training with visually-grounded proxy tasks (\eg, MLM coupled with image and image-sentence matching). UNITER \cite{chen2019uniter}, Unicoder-VL \cite{unicoder}, and VL-BERT \cite{su2019vl} further enhance the region-level vision-language alignment by introducing masked object classification proxy task. Two-stream approaches include ViLBERT \cite{lvilbert} and LXMERT \cite{tan2019lxmert}. These two-stream models typically consist of two separate encoders and one cross-modal encoder. DALL-E \cite{dalle} uses a Transformer model that receives both text and image as a single stream of data and generates images in autoregressively manner. 

Unlike these existing multimodal pre-trained models that only consider two modalities, our OPT jointly models the information of three modalities (\ie vision, language and audio). To the best of our knowledge, this is the first work to learn joint vision+language+audio representation through self-supervised pre-training. 

\section{Model Architecture}
The model architecture of our OPT is illustrated in Figure \ref{fig:model}. First, the model extracts representations of the input text, image and audio using single-modal encoders. Then, a Transformer based cross-modal encoder is used for interacting among the text, the image and the audio. Finally, two cross-modal decoders are used to reconstruct the input text and image, respectively.  

\subsection{Single-Modal Encoders}
\textbf{Text Encoder.} Following BERT \cite{bert}, we first tokenize all words by WordPieces \cite{johnson2017google} to obtain the token sequence $T = \{T_1,...,T_N\}$, where $T_i$ is the $i$-th token and $N$ is the length of the token sequence. The final embedding for each token is obtained via summing up its token embedding and position embedding, followed by an Layer Normalization (LN) layer \cite{ba2016layer}. 

\textbf{Vision Encoder.} We use Faster R-CNN \cite{ren2015faster} pre-trained on Visual Genome dataset \cite{krishna2017visual} to extract the visual representations (pooled ROI features) for each image region. To capture the spatial location information, we introduce a 7-dimensional location feature $[x_1, y_1, x_2, y_2, w, h, w*h]$, where $(x_1, y_1)$ and $(x_2, y_2)$ represent the top left and bottom right coordinates respectively, and $w/h$ is the width/height of the region. Then, both visual and location features are projected into the same embedding space through two fully-connected (FC) layers. The final visual embedding for each region is obtained by summing up the two FC outputs and then passing through an LN layer. 

\textbf{Audio Encoder.} We use pre-trained wav2vec 2.0 \cite{wav2vec} to obtain the audio tokens and extract the features for each token. The final audio embedding is obtained by passing the audio features through an LN layer.

\subsection{Cross-Modal Encoder}
After processing the input text, image and audio using single-modal encoders, we obtain the initial text embedding $\boldsymbol{\mathrm{T}}$, image embedding $\boldsymbol{\mathrm{V}}$ and audio embedding $\boldsymbol{\mathrm{A}}$. To model the cross-modal interations among textual words, visual regions and audio tokens, we introduce a Transformer based cross-modal encoder. Specifically, we first combine $\boldsymbol{\mathrm{T}}$, $\boldsymbol{\mathrm{V}}$ and $\boldsymbol{\mathrm{A}}$ to get a long sequence of token features, and then feed the sequence of features into the cross-modal encoder to learn contextualized representations $\boldsymbol{\mathrm{M}}$,
\begin{align}
\boldsymbol{\mathrm{M}} = \mathrm{CrossEncoder}([\boldsymbol{\mathrm{T}}; \boldsymbol{\mathrm{V}}; \boldsymbol{\mathrm{A}}])
\end{align} 
where $[;]$ denotes the combination operation. Note that the combination is operated along the dimension of sequence, not the dimension of hidden size.

\subsection{Cross-Modal Decoders}  \label{sec:dec}
We empower our pre-trained model to have the capability of learning from and then benefiting for generation tasks by attaching two cross-modal decoders (\ie Text Decoder and Vision Decoder). The Text and Vision Decoder learns to reconstruct the input text and image during pre-training, respectively. When performing on down-stream, the two decoders is used to generate results, \eg, image captioning, text-to-image generation. We adopt Transformer decoder \cite{vaswani2017attention} as our Text Decoder. For Vision Decoder (Figure \ref{fig:img_deco}), following \cite{dalle}, we use a two-stage framework for image generation, including discrete representation learning and language modeling. The first stage focuses on transforming images into sequences of discrete codes. We use dVAE \cite{dalle} for discrete code generation. In the second stage, we build a language model (\ie Transformer decoder) to learn to generate code sequence. We set the objective of autoregressive language modeling for the training. At the stage of inference (\eg, text-to-image generation), we input the text sequence, and the model generates codes autoregressively with top-k sampling. The last step is to transform the code sequence to an image with the dVAE decoder from the first stage.

\section{Pre-Training Tasks}
We propose three levels of pre-training tasks: (1) token-level modeling, including masked language modeling (MLM), masked vision modeling (MVM), and masked audio modeling (MAM); (2) modality-level modeling, including denoising text reconstruction (DTR) and denoising image reconstruction (DIR); and (3) sample-level modeling.

\subsection{Token-Level Modeling}
We introduce token-level modeling task to learn to predict the masked tokens during pre-training. For three different modalities, we propose the corresponding token-level modeling methods, which are described as follows. \\

\noindent{\bfseries Masked Language Modeling (MLM).} We denote the textual words as $T = \{T_1,...,T_N\}$, the image regions as $V = \{V_1,...,V_K\}$, the audio tokens as $A=\{A_1,...,A_Q\}$, and the mask indices as $m$. Following BERT \cite{bert}, we randomly mask 15\% words with the special token $[\mathrm{MASK}]$ in the sentence. The goal is to predict these masked words based on the observation of their surrounding words $T_{\backslash m}$, all image regions $V$ and all audio tokens $A$, by minimizing the negative log-likelihood:
\begin{align}
\mathcal{L}_{\mathrm{MLM}}(\theta) = - \mathbb{E}_{(T,V,A)\backsim D} \mathrm{log}P_\theta (T_m | T_{\backslash m}, V, A)
\end{align}  
where $\theta$ is the trainable parameters. Each pair $(T,V,A)$ is sampled from the whole training set $D$.  \\

\noindent{\bfseries Masked Vision Modeling (MVM).} Similar to MLM, we also propose Masked Vision Modeling (MVM) to predict the correct image regions given contextual regions and other input modalities. We sample image regions and then mask their visual features with a probability of 15\%. The model is trained to reconstruct the masked regions $V_m$ based on the observation of the remaining regions $V_{\backslash m}$, all textual words $T$ and all audio tokens $A$. Different from MLM, the visual features are high-dimensional and continuous, thus it is not feasible to apply the class likelihood objective. Here, we propose two objectives for MVM, which share the same objective base:
\begin{align}
\mathcal{L}_{\mathrm{MVM}}(\theta) = \mathbb{E}_{(T,V,A)\backsim D} f_{\theta}(V_m|T, V_{\backslash m},A)
\end{align}  

The first objective is Masked Visual Feature Regression (MVFR), which regresses the cross-modal encoder output of each masked region $V_m$ to its input ROI visual features $\boldsymbol{\mathrm{V}}_m$. We use an additional FC layer to transform the output of cross-modal encoder to the same dimensional space as the input visual feature. Then we apply L2 regression between the two: 
\begin{align}
f_{\theta}(V_m|T, V_{\backslash m},A) = \sum_m \| h_{\theta}(\boldsymbol{\mathrm{V}}_m) - \boldsymbol{\mathrm{V}}_m\|^2_2
\end{align}
where $h_{\theta}(\cdot)$ denotes the cross-modal encoder plus an FC layer.

The second objective is Masked Region Classification (MRC). MRC learns to predict the object class for each masked region. Due to there is no ground-truth label, we take the detected object category (with the highest confidence score) from Faster RCNN as the label of the masked region. The cross-modal encoder output of each masked region is fed into an FC layer to predict the scores of object classes, which are further transformed into a normalized distribution via a softmax function. The final objective minimizes the cross-entropy (CE) loss:
\begin{align}
 f_{\theta}(V_m|T, V_{\backslash m},A) = \sum_m \mathrm{CE}(g_{\theta}(\boldsymbol{\mathrm{V}}_m), gt(\boldsymbol{\mathrm{V}}_m))
\end{align}
where $g_{\theta}(\cdot)$ consists of the cross-modal encoder, an FC layer and a softmax function, and $gt(\boldsymbol{\mathrm{V}}_m)$ denotes a one-hot vector of ground-truth label. \\

\begin{figure}[t]
\begin{center}
   \includegraphics[width=0.99\linewidth]{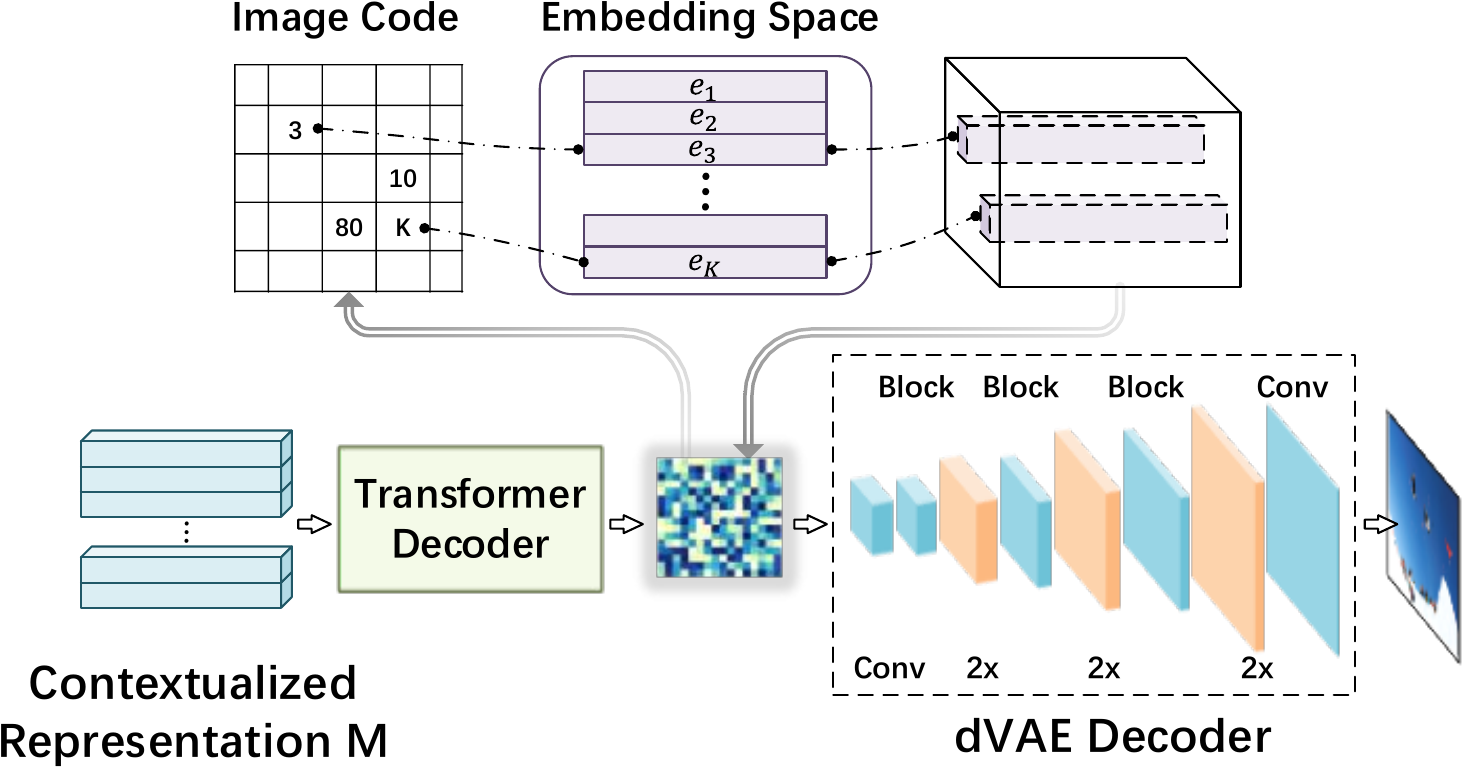}
\end{center}
  \caption{Illustration of Vision Decoder. The vision decoder consists of a Transformer decoder that learns image code and a pre-trained dVAE decoder that generates image.}
\label{fig:img_deco}
\end{figure}

\noindent{\bfseries Masked Audio Modeling (MAM).} For MAM, we mask audio features with a probability of 15\%. Then the model is trained to reconstruct masked audio $A_m$, given the remaining audio tokens $A_{\backslash m}$ and all information from other modalities (\ie text and image). Here, we propose two objectives for MAM, which share the same objective base:
\begin{align}
\mathcal{L}_{\mathrm{MAM}}(\theta) = \mathbb{E}_{(T,V,A)\backsim D} f_{\theta}(A_m|T, V, A_{\backslash m})
\end{align} 

Similar to MVFR, the first objective is Masked Audio Feature Regression (MAFR), which minimizes the L2 regression loss between the input features and the output of masked audio tokens: 
\begin{align}
f_{\theta}(A_m|T, V, A_{\backslash m}) = \sum_m \| h_{\theta}(\boldsymbol{\mathrm{A}}_m) - \boldsymbol{\mathrm{A}}_m\|^2_2
\end{align}

Instead of directly regressing the real values of masked audio features, we adopt the contrastive learning method to maximize the MI (Mutual Information) between the masked output features and the original features. For the output feature of each masked audio token, we pick its original feature to construct the positive pair, and the other tokens as the negative samples. The second objective is defiend as follows: 
\begin{small}
\begin{align}
&f_{\theta}(A_m|T, V, A_{\backslash m}) = \notag \\
&-\mathrm{log} \frac{\mathrm{exp}(sim(h_\theta(\boldsymbol{\mathrm{A}}_m), \boldsymbol{\mathrm{A}}_m))}{\mathrm{exp}(sim(h_\theta(\boldsymbol{\mathrm{A}}_m), \boldsymbol{\mathrm{A}}_m)) + \mathrm{exp}(sim(h_\theta(\boldsymbol{\mathrm{A}}_m), \boldsymbol{\mathrm{A}}_{\backslash m}))}
\end{align}
\end{small}
where $sim(\cdot,\cdot)$ is the cosine similarity, and $h_\theta(\cdot)$ includes the cross-modal encoder and an FC layer as in other pre-training tasks metioned above.

\subsection{Modality-Level Modeling}
To endow the pre-trained model with the generation capability and also further benefit the representation learning, we propose modality-level modeling task along with modality-level masking mechanism to reconstruct one whole modality. The modality-level modeling tasks includes text reconstruction and image reconstruction, which are described as follows.  \\

\noindent{\bfseries Modality-Level Masking.} We propose modality-level masking mechanism to learn the alignment amiong the three modalities, \ie text, vision, and audio. Modality-level masking is in parallel with token-level masking mechanism. It masks out one or two modalities from the input. specifically, each modality is independently masked out with a probability of 0.3, and the case when all modalities are masked is skipped. This brings a significant benefit  $-$ allowing OPT to handle different downstream tasks with single-, two-, or three-modality inputs. \\

\noindent{\bfseries Denoising Text Reconstruction (DTR).} We attach a Transformer based decoder \cite{vaswani2017attention} on the top of the cross-modal encoder to learn to reconstruct the input text. The loss function is,
\begin{align}
\mathcal{L}_{\mathrm{DTR}}(\theta) = - \mathbb{E}_{(T,V,A) \backsim D} \mathrm{log} P_\theta (\hat{T}_i | \hat{T}_{<i}, T, V, A)    \label{tr_loss}
\end{align} 
\\
\noindent{\bfseries Denoising Image Reconstruction (DIR).} We also employ an vision decoder to endow our OPT model with the capability of image generation. The decoder is trained to learn to reconstruct the input image. As shown in Figure \ref{fig:img_deco}, the vision decoder consists of the Transformer decoder and the dVAE decoder. The Transformer decoder generates a sequence of image codes. We enforce an autoregressive language modeling loss as in Eq. \ref{tr_loss} to learn good image codes. The pre-trained dVAE decoder from the first stage (see Sec. \ref{sec:dec}) is frozen during our OPT pre-training. The loss function is as follows: 
\begin{align}
\mathcal{L}_{\mathrm{DIR}}(\theta) = - \mathbb{E}_{(T,V,A) \backsim D} \mathrm{log} P_\theta (\hat{I}_i | \hat{I}_{<i}, T, V, A) 
\end{align}
 
\subsection{Sample-Level Modeling}
We design a sample-level modeling task for three-modality pre-training. Specifically, given each sample (\ie text-image-audio triplet), we randomly replace one or two inputs with ones from other samples. The model requires to predict which inputs are matching. As a result, we have five cases: (1) all three inputs are matching; (2) only image and audio are matching; (3) only text and image are matching; (4) only text and audio are matching; and (5) no one is matching, as shown in Figure \ref{fig:model} (the sample-level modeling task). We extract the output representation of $[\mathrm{CLS}]$ token as the joint representation of the text-image-audio triplet, then feed it into an FC layer and a sigmoid function to predict the scores. We denote the output scores as $s_\theta(T,I,A) \in \mathbb{R}^5$. The loss function is the binary cross-entropy (BCE) loss:
\begin{align}
\mathcal{L}_{\mathrm{SM}}(\theta) = \mathbb{E}_{(T,V,A)\backsim D}\mathrm{BCE}(s_\theta(T,V,A), gt(T,V,A)) 
\end{align}  
where $gt(T,V,A)$ is the one-hot vector of ground-truth label.

\begin{table}[!t]
\begin{center}
\caption{Statistics of some common pre-training datasets.}
\label{tab:data}
\resizebox{8.2cm}{!}{
\begin{tabular}{l | c c c c c}
\toprule[1pt]
Datasets & \#images & \#captions & \#audios   \\
\midrule
COCO Captions (train) & 0.1M & 0.5M & -  \\
Flickr30k (train) & 29K & 0.1M & - \\
VQA (train) & 83K & 0.4M & -  \\
GQA (train) & 82K & 1.1M & -  \\
VG & 0.1M & 5.1M & - \\
Conceptual Captions & 3.1M & 3.1M & -  \\
SBU Captions & 1.0M & 1.0M & - \\
Open Images & 0.6M & 0.6M & 0.6M  \\
\bottomrule[1pt]
\end{tabular}
}
\end{center}
\end{table}

\section{Experiments}
In this section, we conduct comprehensive experiments on downstream tasks and provide ablation studies to validate the effectiveness of different pre-training settings. 

\subsection{Pre-Training Dataset}
We mainly use Open Images \cite{kuznetsova2020open} dataset with localized narratives and synchronized speech that are provided by \cite{pont2020connecting} as our pre-training dataset. Only text-image-audio triplets are used for pre-training. The dataset consists of 641,716 images with captions and speeches. We exclude all the images that appear in the downstream tasks to avoid contamination in evaluation. We randomly select 5,000 samples from Open Images test dataset as our test set for downstream ablative experiments, and name it OpenImages-5K. When comparing with state-of-the-art methods on downstream two-modal tasks, we also add commonly-used two-modal datasets (\eg Conceptual Captions \cite{sharma2018conceptual}, VG \cite{krishna2017visual}) as our pre-training datasets. Table \ref{tab:data} shows the data volume of some commonly-used pre-training datasets. 

\subsection{Implementation Details}
We use Faster-RCNN \cite{ren2015faster} (with ResNet-152 backbone) pretrained on the Visual Genome dataset \cite{krishna2017visual} to extract image region features. We select regions where class confidence exceeds a threshold and keep between 10 to 100 high-scoring boxes. We apply pre-trained wave2vec 2.0 \cite{wav2vec} framework to tokenize the input audio and extract the audio token features. For the cross-modal encoder, we use the BERT-base model \cite{bert} with 12 layers of Transformer blocks. Each block has 12 attention heads and the hidden size is 768. The text decoder and the Transformer decoder in the vision decoder has 6 layers of blocks. The size of reconstructed image is 64$\times$64. The size of image code is 8$\times$8, and the embedding dimensionality is 8192. Models are trained on 4 Tesla V100 GPUs with a total batch size of 10,240 for 100,000 iterations, and early stop is performed. We adopt the Adam optimizer with an initial learning rate of 5e-5. 

\begin{table}[!t]
\begin{center}
\caption{Linear probe results on multi-label classification task. The performance is evaluated on Open Images val set.}
\label{tab:mulcls}
\resizebox{7.2cm}{!}{
\begin{tabular}{l | c c c | c}
\toprule[1pt]
Method & Text & Image & Audio & mAP   \\
\midrule
ResNet-50 \cite{he2016deep} & & $\surd$ & & 52.20  \\
ResNet-101 \cite{he2016deep} & & $\surd$ & & 53.10 \\
\midrule
OPT (ours) & $\surd$ & & & 49.20 \\
OPT (ours) & & $\surd$ & & 56.00   \\
OPT (ours) & & & $\surd$ & 53.84  \\
\midrule
OPT (ours) & $\surd$ & $\surd$ & & 57.86  \\
OPT (ours) & $\surd$ & & $\surd$ & 54.00   \\
OPT (ours) & & $\surd$ & $\surd$ & 56.59  \\
\midrule
OPT (ours) & $\surd$ & $\surd$ & $\surd$ & \textbf{58.11}  \\
\bottomrule[1pt]
\end{tabular}
}
\end{center}
\end{table}

\subsection{Results}
We qualitatively and quantitatively validate the effectiveness of our proposed model on both understanding tasks (including classification, retrieval, etc.) and generation tasks (including text generation and image generation).

\textbf{Multi-Modal Classification.} We first conduct experiments on multi-modal classification task. We add a linear layer after the average pooling output of cross-modal encoder for classification. We freeze our pre-trained model and only linear layer is learned. The experimental results are shown in Table \ref{tab:mulcls}. When only using image features, our OPT outperforms ResNet-50 and ResNet-101 by a large margin. When only using text or audio feature, our model also obtains promising results, which indicates that the model has learnt the associations between different modalities. When with multimodal features, the performance can be further improved. In particular, adding text feature brings the largest improvement. On the basis of image+text feature, we further add audio feature, and find the performance still increases. This shows the benefit of multi-modal inputs, each of which contributes to the model performance.  

\begin{table}[!t]
\begin{center}
\caption{Results on cross-modal retrieval task. The performance is evaluated on OpenImages-5K test set. ``A $\to$ B'' means using A to retrieve B.}
\label{tab:mulret}
\resizebox{8.4cm}{!}{
\begin{tabular}{l | l | c c c}
\toprule[1pt]
Retrieval Task & Method & R@1 & R@5 & R@10     \\
\midrule
 & Random & 0.02 & 0.10 & 0.20 \\
\midrule
\multirow{2}*{Image $\to$ Text} & ViLBERT \cite{lvilbert} & 12.72 & 30.84 & 38.96   \\
 & OPT (ours) & 39.40 & 71.94 & 82.56   \\
\midrule
\multirow{2}*{Text $\to$ Image} & ViLBERT \cite{lvilbert} & 0.00 & 26.66 & 38.96  \\
 & OPT (ours) & 41.96 & 72.00 & 81.26   \\  
\midrule
Text $\to$ Audio & OPT (ours) & 78.00 & 92.70 & 95.80  \\
\midrule
Audio $\to$ Text & OPT (ours) & 80.30 & 94.50 & 97.10  \\
\midrule
Text-Audio $\to$ Image & OPT (ours) & 57.06 & 79.04 & 85.78  \\
\bottomrule
\end{tabular}
}
\end{center}
\end{table}

\begin{table}[!t]
\begin{center}
\caption{Results on text generation for audio recognition task. The performance is evaluated on OpenImages-5K test set.}
\label{tab:sr}
\resizebox{8.2cm}{!}{
\begin{tabular}{l |c c c | c}
\toprule[1pt]
Method & Text & Image & Audio & Word Error Rate (WER)  \\
\midrule
ESPnet \cite{espnet} &  & & $\surd$ & 46.89  \\
Baidu API\footnote{\url{https://ai.baidu.com/ai-doc/SPEECH/Vk38lxily}} & & & $\surd$ & 48.35  \\
IBM API\footnote{\url{https://cloud.ibm.com/apidocs/speech-to-text}} & & & $\surd$ & 57.47  \\
\midrule
OPT (ours) & & & $\surd$ & 31.35   \\
OPT (ours) & & $\surd$ & $\surd$ & \textbf{30.24}   \\
\bottomrule[1pt]
\end{tabular}
}
\end{center}
\end{table}

\textbf{Cross-Modal Retrieval.} We evaluate our OPT model on different cross-modal retrieval tasks on OpenImages-5K test set. We select several typical retrieval tasks, including text-image retrieval, text-audio retrieval, image-text retrieval, audio-text retrieval. It can be seen that our OPT outperforms ViLBERT \cite{lvilbert} by a large margin on Image-Text and Text-Image retrieval. Note that since ViLBERT only uses vision and text information, we are unable to test it on other retrieval tasks involing audio. We also evaluate our model on the retrieval tasks using more modalities \ie, we use the information of both modalities to retrieve the rest one modality. We find that additional input modality can significantly improve the retrieval performance (\eg Text $\to$ Image \textit{v.s.} Text-Audio $\to$ Image). Thus, it is necessary to jointly model the information of more modalities. 

\begin{table}[!t]
\begin{center}
\caption{Ablation studies of our model on multi-modal classification task.}
\label{tab:as}
\resizebox{6cm}{!}{
\begin{tabular}{l | c}
\toprule[1pt]
Setting & mAP   \\
\midrule
w/o Masked Language Modeling & 53.02    \\
w/o Masked Vision Modeling & 52.62   \\
w/o Masked Audio Modeling & 54.24   \\
w/o Modality-Level Modeling & 51.01  \\
w/o Sample-Level Modeling & 50.72   \\
\midrule
Full & 58.11     \\
\bottomrule[1pt]
\end{tabular}
}
\end{center}
\end{table}

\begin{table*}[!t]
\begin{center}
\caption{Performance comparison with previous methods on two-modal (\ie visual-text) downstream tasks. CC: Conceptual Captions. COCO: COCO Captions. VG: Visual Genome dataset. SBU: SBU Captions. OI: Open Images. The data volume of each pre-training dataset is shown in Table \ref{tab:data}.}
\label{tab:sota}
\begin{tabular}{l c c c c c}
\toprule[1pt]
\multirow{2}*{Method}& \multirow{2}*{Pre-training Datasets} & Flickr-IR & Flickr-TR & VQA &COCO-Caption  \\
 & & R@1/R@5/R@10 & R@1/R@5/R@10 & Test-dev/std & BLEU4/CIDEr   \\
\midrule
SOTA & None & 48.60/77.70/85.20 & 67.90/90.30/95.80 & 70.63/70.90 & 37.2/119.8  \\
\midrule
ViLBERT \cite{lvilbert} & CC & 58.20/84.90/91.52 & - & 70.55/70.92 & -   \\
LXMERT \cite{tan2019lxmert} & COCO+VG+VQA+GQA & - & - & 72.42/72.54 & - \\
VLP \cite{zhou2020unified}& CC & - & - & 70.50/70.70 & 36.5/116.9   \\
UNITER \cite{chen2019uniter}& COCO+VG+CC+SBU & 72.52/92.36/96.08 & 85.90/97.10/\textbf{98.80} & 72.70/72.91 & -   \\
\multirow{2}*{Oscar \cite{li2020oscar}} & COCO+CC+SBU+Flickr & \multirow{2}*{-} & \multirow{2}*{-} & \multirow{2}*{\textbf{73.16}/\textbf{73.44}} & \multirow{2}*{36.5/123.7}  \\
 & +VQA+GQA &   \\
\midrule
OPT (ours) & OI & 64.06/87.32/92.34 & 79.70/95.10/97.60 & 71.70/72.02 & 39.1/129.5   \\
OPT (ours) & OI+VG+CC & \textbf{73.58}/\textbf{92.60}/\textbf{96.54} & \textbf{86.20}/\textbf{97.50}/98.60 & 72.38/72.64 & \textbf{40.2}/\textbf{133.8}  \\
\bottomrule[1pt]
\end{tabular}
\end{center}
\end{table*}

\textbf{Text Generation for Audio Recognition.} In Table \ref{tab:sr}, we test our model on audio recognition task. With audio only or both audio and image as inputs, we calculate the word error rate (WER) between model's output text and ground truth. The compared methods include several API from Baidu and IBM companies and a state of the art model, Espnet \cite{espnet}, which is pretrained on the librispeech dataset, resulting 48.35, 57.47 and 46.89 WER respectively. These methods take audio as input and all of them are tested on the same OpenImages-5K dataset. It can be seen that OPT outperforms these compared methods by a large margin, improving at least 15 point. In particular, with image feature, the performance of audio recognition can be further improved about 1 point.

\textbf{Performance on Visual-Text Downstream Tasks.} Since there is no good benchmarks and pre-training methods based on three modalities, we compare with state-of-the-art methods on the benchmarks based on two modalities. Table \ref{tab:sota} shows the performance comparison. With Open Images as pre-training dataset (only 0.6M samples), our model outperforms other two-modal methods which use much more pre-training samples on COCO-Caption task. When add more image-text pairs as pre-training datasets, the performance of our model is significantly improved, and surpasses other methods on most tasks. 

\subsection{Ablation Study}
We conduct ablation studies on multi-modal classification task. We respectively remove each pre-training task to investigate the effect of each component. The results are shown in Table \ref{tab:as}. Ablating each pre-training task leads to severe performance drop. Specifically, ``w/o Sample-Level Modeling'' results in the largest performance drop (around 7.4\%). This is because this pre-training task can enforce stronger and more fine-grained correlations between multiple modalities. ``w/o Modality-Level Modeling'' produces $\thicksim$7\% performance drop, showing that incorporating generation task benefits the representation learning. For token-level modeling, masked vision modeling contributes the most to the final performance, possibly due to the evaluation task is multi-modal classification that heavily relies on visual information. 

\subsection{Qualitative Results}
\begin{figure*}[t]
\begin{center}
   \includegraphics[width=0.99\linewidth]{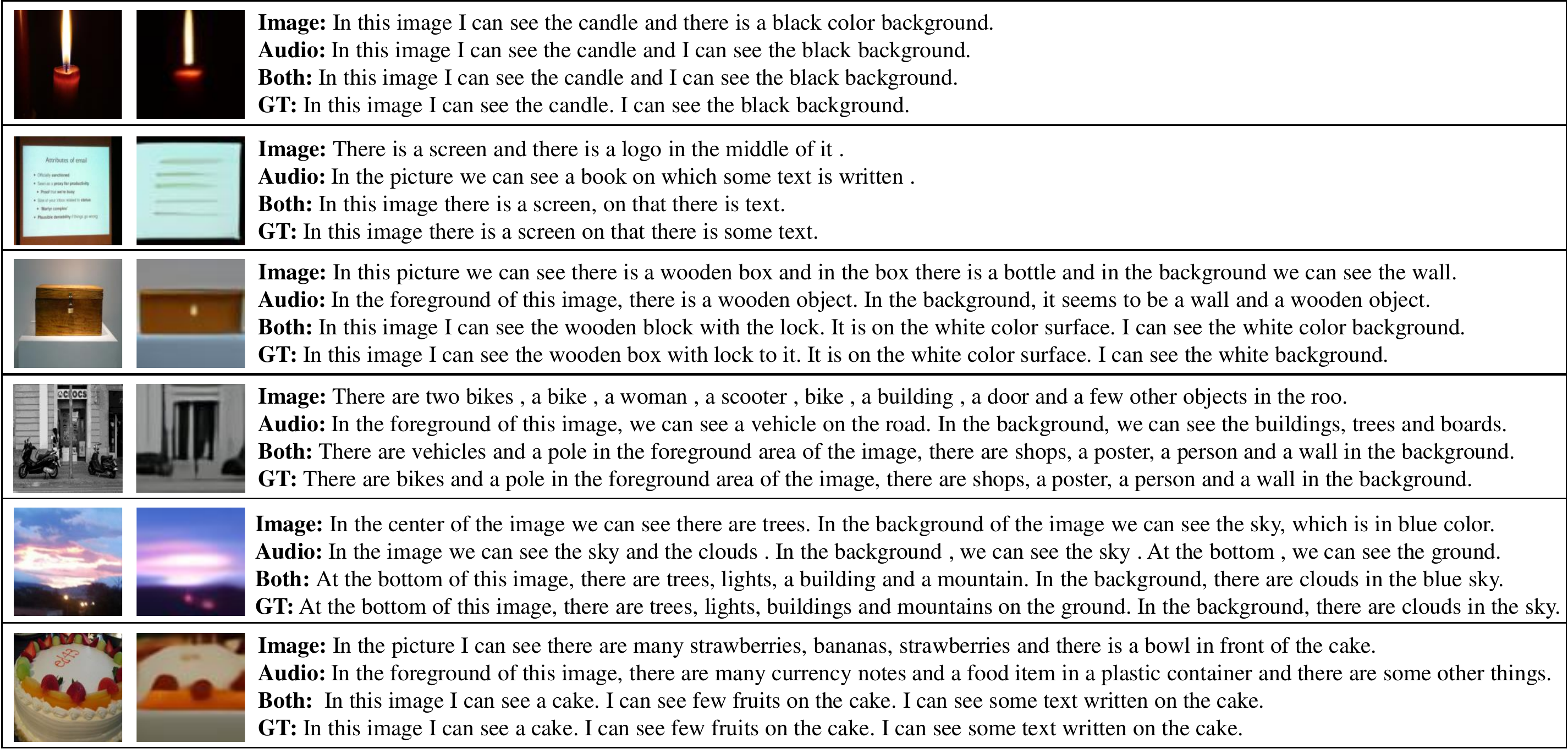}
\end{center}
  \caption{Some results of text-to-image generation and text generation (including \textbf{Image:} image captioning, \textbf{Audio:} audio recognition, and \textbf{Both:} text generation with image+audio). The 1st column shows the ground-truth images and the 2nd column shows the generated images.}
\label{fig:vis}
\end{figure*}

In Figure \ref{fig:vis}, we provide some generated results from our model. The first two columns show the results of text-to-image generation. Our model can learn some specific patterns and generally reconstruct the image. On the right side of Figure \ref{fig:vis}, we show the corresponding text generation, including imaging captioning, audio recognition, and the text generation with image+audio. It can be seen that our model is able to generate very accurate sentences. These results prove that image generation and text generation can be integrated into a unified framework. It is noted that we make a first attempt to incorporate the image generation into the pre-trained model. There is still room to improve the image decoder and image reconstruction pre-training task, which we leave for future work.

\section{Discussion}
OPT has taken the first step towards cross-modal understanding and generation on text, vision, and audio modalities, and verified the feasibility and effectiveness of the unified pre-training solution for omni-modality perception. There remain many open problems to be solved. Compared with single-modality or two-modalities resources, three modality triplets for pretraining is more difficult to collect. How to efficiently train the model under un-paired  or partial-modality data (randomly or always misses one or two modalities) is conducive to enhance the robustness and generation of the model. In order to make the model have the human-like reasoning ability, it is also necessary to introduce knowledge modeling into the pre-training process. Besides, we can attempt to learn from human interaction or feedback (available on the Internet at low cost) with techniques like reinforcement learning etc. 
 
In addition to the above improvements for the model pre-training, many interesting applications could be explored on OPT-like models. For example, empowering the model with the ability to generate raw audio (conditioned on language, image, or another audio), image editing and image to image translation, video editing and generation under language/audio instructions etc. 

\section{Conclusion}
In this paper, we present an Omni-perception Pre-Trainer (OPT) for cross-modal understanding and generation, by jointly modeling visual, text and audio resources. OPT follows an encoder-decoder framework. For the OPT's pre-training, we design multi-level pre-training tasks, including token-, modality- and sample-level modeling. Experimental results on a number of downstream tasks verify the effectiveness of our OPT. In the future, we plan to include the audio generation into our framework. 

{\small
\bibliographystyle{ieee_fullname}
\bibliography{egpaper_for_review}
}

\end{document}